\documentclass[11pt,a4paper]{article}
\usepackage[hyperref]{emnlp-ijcnlp-2019}
\usepackage{times}
\usepackage{latexsym}
\usepackage{graphicx}
\usepackage{tabularx}
\usepackage{wrapfig}
\usepackage[normalem]{ulem}
\usepackage{amsmath}
\usepackage{subcaption}
\usepackage{xcolor}
\usepackage{multirow}
\usepackage{multicol}
\usepackage{makecell}
\usepackage{amssymb}
\def\squiggly{\bgroup \markoverwith{\textcolor{red}{\lower3.5\p@\hbox{\sixly \char58}}}\ULon}

\usepackage{url}
\usepackage{xcolor} 

\usepackage{paralist}

\aclfinalcopy

\title{Rethinking Attribute Representation and Injection\\ for Sentiment Classification}

\author{Reinald Kim Amplayo \\
  Institute for Language, Cognition and Computation \\
  School of Informatics, University of Edinburgh \\
  {\tt reinald.kim@ed.ac.uk} \\
}
\date{}

\makeatletter
\newcommand{\thickhline}{%
    \noalign {\ifnum 0=`}\fi \hrule height 1pt
    \futurelet \reserved@a \@xhline
}
\makeatother

\begin{document}
\maketitle
\begin{abstract}
  Text attributes, such as user and product information in product reviews, have been used to improve the performance of sentiment classification models. The de facto standard method is to incorporate them as additional biases in the attention mechanism, and more performance gains are achieved by extending the model architecture. In this paper, we show that the above method is the least effective way to represent and inject attributes. To demonstrate this hypothesis, unlike previous models with complicated architectures, we limit our base model to a simple BiLSTM with attention classifier, and instead focus on \textit{how} and \textit{where} the attributes should be incorporated in the model. We propose to represent attributes as chunk-wise importance weight matrices and consider four locations in the model (i.e., embedding, encoding, attention, classifier) to inject attributes. Experiments show that our proposed method achieves significant improvements over the standard approach and that attention mechanism is the worst location to inject attributes, contradicting prior work. We also outperform the state-of-the-art despite our use of a simple base model. Finally, we show that these representations transfer well to other tasks\footnote{ Model implementation and datasets are released here: \url{https://github.com/rktamplayo/CHIM}}.
\end{abstract}

\section{Introduction}

The use of categorical attributes (e.g., user, topic, aspects) in the sentiment analysis community \cite{kim2004determining,pang2007opinion,liu2012sentiment} is widespread.
Prior to the deep learning era, these information were used as effective categorical features \cite{li2011incorporating,tan2011user,gao2013modeling,park2015retrieval} for the machine learning model. Recent work has used them to improve the overall performance \cite{chen2016neural,dong2017learning}, interpretability \cite{amplayo2018cold,angelidis2018summarizing}, and personalization \cite{ficler2017controlling} of neural network models in different tasks such as sentiment classification \cite{tang2015learning}, review summarization \cite{yang2018cross}, and text generation \cite{dong2017learning}.

In particular, user and product information have been widely incorporated in sentiment classification models, especially since they are important metadata attributes found in review websites. \citet{tang2015learning} first showed significant accuracy increase of neural models when these information are used. Currently, the accepted standard method is to use them as additional biases when computing the weights $a$ in the attention mechanism, as introduced by \citet{chen2016neural} as:
\begin{align}
    a &= \text{softmax}(v^\top e) \\
    \label{eq:upa}
    e &= \tanh (Wh + W_u u + W_p p + b) \\
      &= \tanh (Wh + b_u + b_p + b) \\
      &= \tanh (Wh + b')
\end{align}
where $u$ and $p$ are the user and product embeddings, and $h$ is a word encoding from BiLSTM. Since then, most of the subsequent work attempted to improve the model by extending the model architecture to be able to utilize external features \cite{zhu2017parallel}, handle cold-start entities \cite{amplayo2018cold}, and represent user and product separately \cite{ma2017cascading}.

Intuitively, however, this method is not the ideal method to represent and inject attributes because of two reasons.
First, representing attributes as additional biases cannot model the relationship between the text and attributes. Rather, it only adds a user- and product-specific biases that are independent from the text when calculating the attention weights.
Second, injecting the attributes in the attention mechanism means that user and product information are only used to customize how the model choose which words to focus on, as also shown empirically in previous work \cite{chen2016neural,ma2017cascading}. However, we argue that there are more intuitive locations to inject the attributes such as when contextualizing words to modify their sentiment intensity.

We propose to represent user and product information as weight matrices (i.e., $W$ in the equation above). Directly incorporating these attributes into $W$ leads to large increase in parameters and subsequently makes the model difficult to optimize. To mitigate these problems, we introduce chunk-wise importance weight matrices, which (1) uses a weight matrix smaller than $W$ by a chunk size factor, and (2) transforms these matrix into gates such that it corresponds to the relative importance of each neuron in $W$. We investigate the use of this method when injected to several locations in the base model: word embeddings, BiLSTM encoder, attention mechanism, and logistic classifier.

The results of our experiments can be summarized in three statements. First, our preliminary experiments show that doing bias-based attribute representation and attention-based injection is not an effective method to incorporate user and product information in sentiment classification models. Second, despite using only a simple BiLSTM with attention classifier, we significantly outperform previous state-of-the-art models that use more complicated architectures (e.g., models that use hierarchical models, external memory networks, etc.). Finally, we show that these attribute representations transfer well to other tasks such as product category classification and review headline generation.

\section{How and Where to Inject Attributes?}
\label{sec:models}

In this section, we explore different ways on how to represent attributes and where in the model can we inject them.

\subsection{The Base Model}

The majority of this paper uses a base model that accepts a review $\mathbf{x}=x_1,...,x_n$ as input and returns a sentiment $y$ as output, which we extend to also accept the corresponding user $u$ and product $p$ attributes as additional inputs.
Different from previous work where models use complex architectures such as hierarchical LSTMs \cite{chen2016neural,zhu2017parallel} and external memory networks \cite{dou2017capturing,long2018dual}, we aim to achieve improvements by only modifying how we represent and inject attributes. Thus, we use a simple classifier as our base model, which consists of four parts explained briefly as follows.

First, we embed $\mathbf{x}$ using a word embedding matrix that returns word embeddings $x'_1,...,x'_n$. We subsequently apply a non-linear function to each word:
\begin{equation}
    w_t = \tanh (W_\text{emb} x'_t + b_\text{emb})
\end{equation}
Second, we run a bidirectional LSTM \cite{hochreiter1997long} encoder to contextualize the words into $h_t=[\overrightarrow{h}_t;\overleftarrow{h}_t]$ based on their forward and backward neighbors. The forward and backward LSTM look similar, thus for brevity we only show the forward LSTM below:
\begin{align}
    \begin{bmatrix} 
    g_t \\ i_t \\ f_t \\ o_t
    \end{bmatrix}
    &= 
    \begin{bmatrix} \tanh \\ \sigma \\ \sigma \\ \sigma
    \end{bmatrix}
    W_\text{enc} [w_t; \overrightarrow{h}_{t-1}] + b_\text{enc} \\
    c_t &= f_t * c_{t-1} + i_t * g_t \\
    \overrightarrow{h}_t &= o_t * c_t
\end{align}
Third, we pool the encodings $h_t$ into one document encoding $d$ using attention mechanism \cite{bahdanau2015neural}, where $v$ is a latent representation of informativeness \cite{yang2016hierarchical}:
\begin{align}
    e_t &= \tanh (W_\text{att} h_t + b_\text{att}) \\
    a_t &= \text{softmax}_t (v^\top e_t) \\
    d &= \sum_t (a_t * h_t)
\end{align}
Finally, we classify the document using a logistic classifier to get a predicted $y'$:
\begin{equation}
    y' = \text{argmax} (W_\text{cls} d + b_\text{cls})
\end{equation}
Training is done normally by minimizing the cross entropy loss.

\subsection{How: Attribute Representation}

\begin{figure}[t]
    \centering
    \begin{subfigure}{\columnwidth}
        \includegraphics[width=\columnwidth]{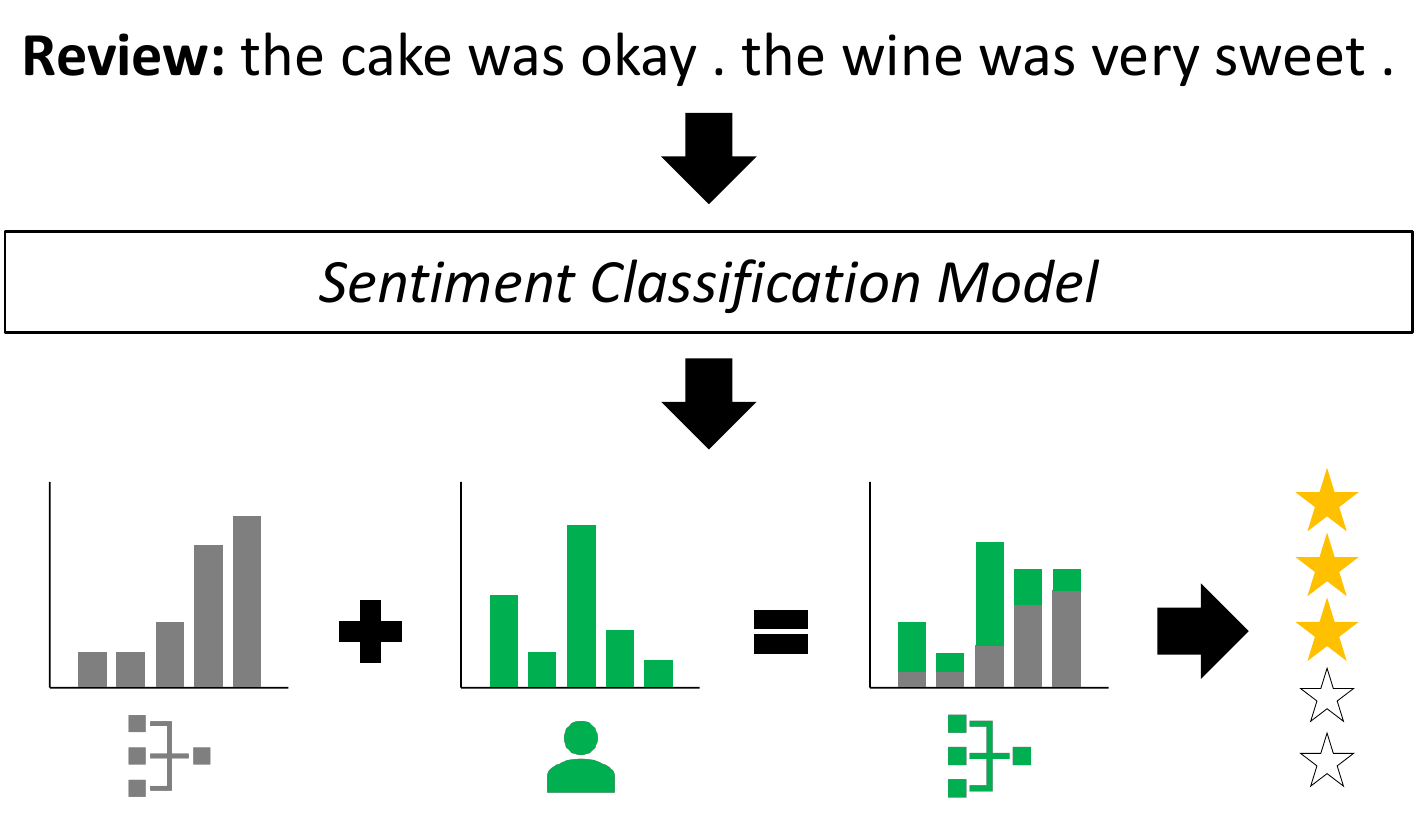}
        \caption{Logits when representing attributes as biases in the logistic classifier}
        \label{fig:bias}
    \end{subfigure}
    \vspace*{0.2em}
    \begin{subfigure}{\columnwidth}
        \includegraphics[width=\columnwidth]{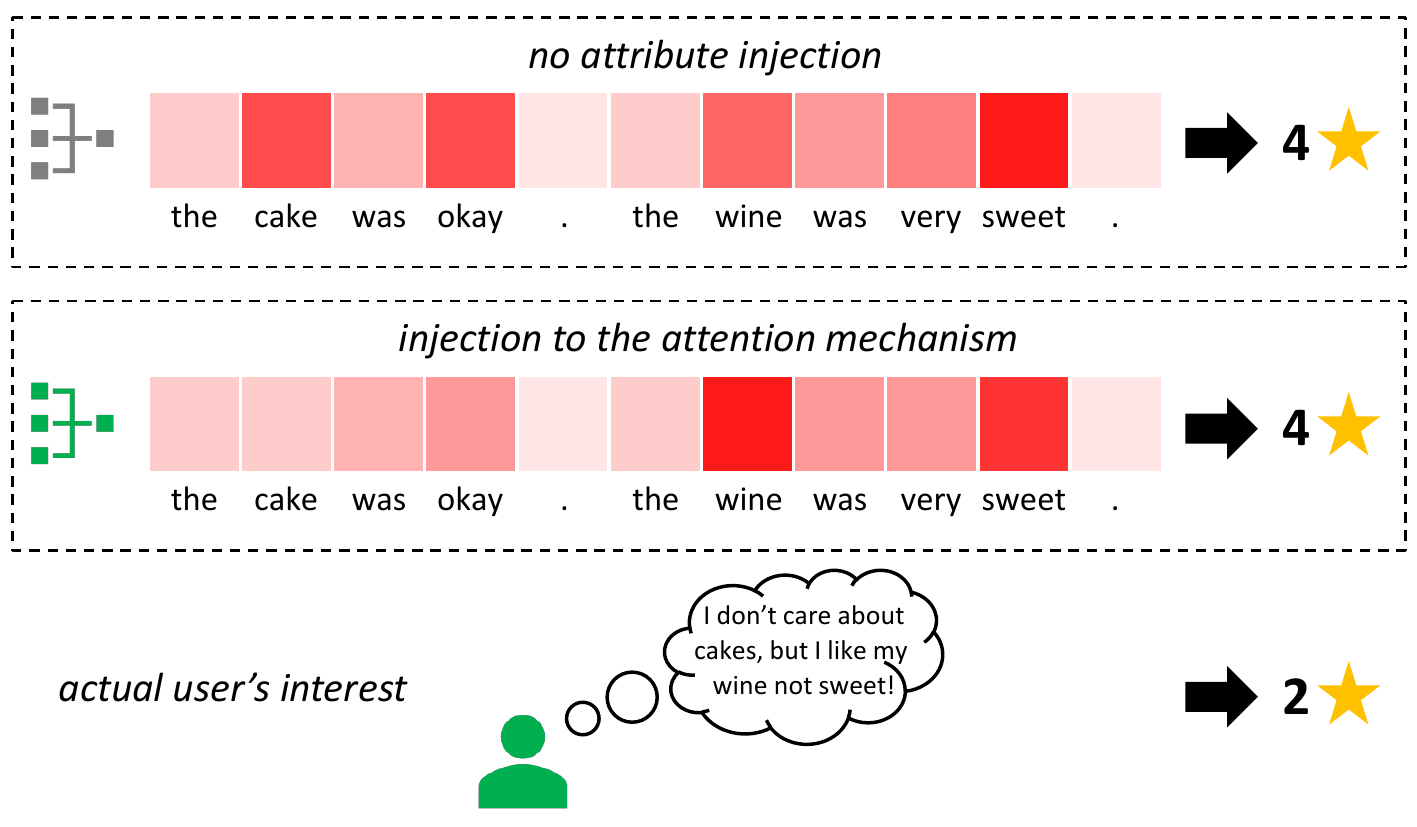}
        \caption{Attention weights when injecting attributes in the attention mechanism.}
        \label{fig:att}
    \end{subfigure}
    \caption{Illustrative examples of issues when representing attributes as biases and injecting them in the attention mechanism. The gray process icon indicates the model without incorporating attributes, while the same icon in green indicates the model customized for the green user.}
\end{figure}

Note that at each part of the model, we see similar non-linear functions, all using the same form, i.e. $g(f(x)) = g(Wx + b)$, where $f(x)$ is an affine transformation function of $x$, $g$ is a non-linear activation, $W$ and $b$ are weight matrix and bias parameters, respectively. Without extending the base model architecture, we can represent the attributes either as the weight matrix $W$ or as the bias $b$ to one of these functions by modifying them to accept $u$ and $p$ as inputs, i.e. $f(x,u,p)$.

\paragraph{Bias-based}

The current accepted standard approach to represent the attributes is through the bias parameter $b$. Most of the previous work \cite{chen2016neural,zhu2017parallel,amplayo2018cold,wu2018improving} use Equation \ref{eq:upa} in the attention mechanism, which basically updates the original bias $b$ to $b' = W_u u + W_p p + b$. However, we argue that this is not the ideal way to incorporate attributes since it means we only add a user- and product-specific bias towards the goal of the function, \textit{without} looking at the text. 
Figure \ref{fig:bias} shows an intuitive example: When we represent user $u$ as a bias in the logistic classifier, in which it means that $u$ has a biased logits vector $b_u$ of classifying the text as a certain sentiment (e.g., $u$ tends to classify texts as three-star positive), shifting the final probability distribution regardless of what the text content may have been.

\paragraph{Matrix-based}

A more intuitive way of representing attributes is through the weight matrix $W$. Specifically, given the attribute embeddings $u$ and $p$, we linearly transform their concatenation into a vector $w'$ of size $D_1*D_2$ where $D_1$ and $D_2$ are the dimensions of $W$. We then reshape $w'$ into $W'$ to get the same shape as $W$ and replace $W$ with $W'$:
\begin{align}
    \label{eq:attribute_transform}
    w' &= W_c [u;p] + b_c \\
    \label{eq:reshape}
    W' &= \text{reshape} (w', (D_1 \times D_2)) \\
    f(x,u,p) &= W'x + b
\end{align}
Theoretically, this should perform better than bias-based representations since direct relationship between text and attributes are modeled. For example, following the example above, $W'x$ is a user-biased logits vector based on the document encoding $d$ (e.g., $u$ tends to classify texts as two-star positive when the text mentions that the dessert was sweet).

However, the model is burdened by a large number of parameters; matrix-based attribute representation increases the number of parameters by $|U|*|P|*D_1*D_2$, where $|U|$ and $|P|$ correspond to the number of users and products, respectively. This
subsequently makes the weights difficult to optimize during training. Thus, directly incorporating attributes into the weight matrix may cause harm in the performance of the model.

\paragraph{CHIM-based}

\begin{figure}[t]
    \centering
    \includegraphics[width=0.9\columnwidth]{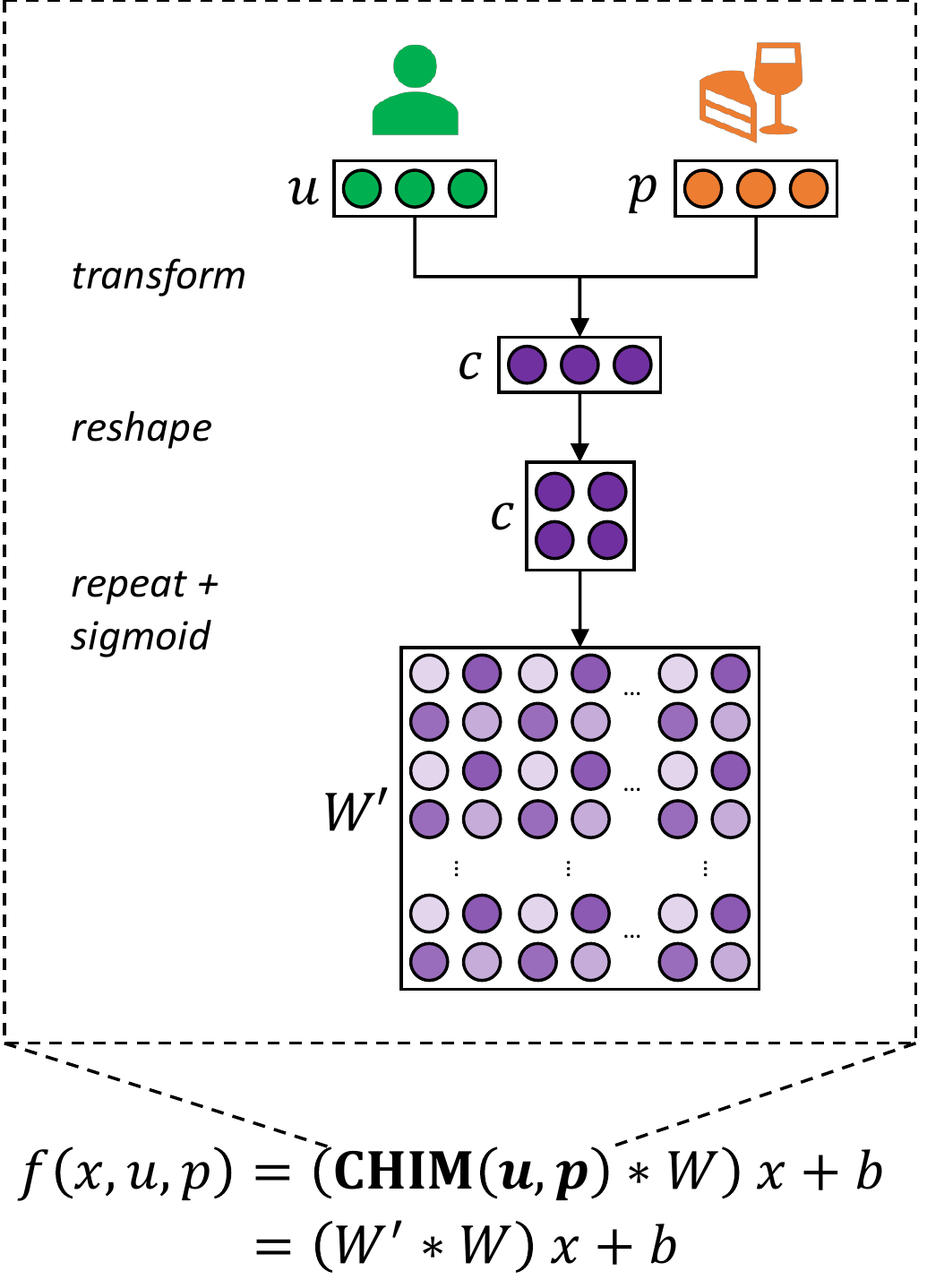}
    \caption{CHIM-based attribute representation and injection to a non-linear funtion in the model.}
    \label{fig:model}
\end{figure}

We introduce \textbf{Ch}unk-wise \textbf{I}mportance \textbf{M}atrix (CHIM) based representation, which improves over the matrix-based approach by mitigating the optimization problems mentioned above, using the following two tricks. First, instead of using a big weight matrix $W'$ of shape $(D_1, D_2)$, we use a chunked weight matrix $C$ of shape $(D_1/C_1, D_2/C_2)$ where $C_1$ and $C_2$ are chunk size factors. Second, we use the chunked weight matrix as importance gates that shrinks the weights close to zero when they are deemed unimportant. We show the CHIM-based representation method in Figure \ref{fig:model}.

We start by linearly transforming the concatenated attributes into $c$. Then we reshape $c$ into $C$ with shape $(D_1/C_1, D_2/C_2)$. These operations are similar to Equations \ref{eq:attribute_transform} and \ref{eq:reshape}. We then repeat this matrix $C_1*C_2$ times and concatenate them such that we create a matrix $W'$ of shape $(D_1, D_2)$. Finally, we use the sigmoid function $\sigma$ to transform the matrix into gates that represent importance:
\begin{equation}
    W'= \sigma
    \begin{bmatrix} 
    C; ...; C \\ ...; ...; ... \\ C; ...; C
    \end{bmatrix}
    \in [0,1]^{D_1 \times D_2}
\end{equation}
Finally we broadcast-multiply $W'$ with the original weight matrix $W$ to shrink the weights. The result is a sparse version of $W$, which can be seen as either a regularization step \cite{ng2004feature} where most weights are set close to zero, or a correction step \cite{amplayo2018translations} where the important gates are used to correct the weights. 
The use of multiple chunks regards CHIM as coarse-grained access control \cite{shen2019ordered} where the use of different important gates for every node is unnecessary and expensive.
The final function is shown below:
\begin{equation}
    f(x,u,p) = (W'*W) x + b
\end{equation}

To summarize, chunking helps reduce the number of parameters while retaining the model performance, and importance matrix makes optimization easier during training, resulting to a performance improvement.
We also tried alternative methods for importance matrix such as residual addition
(i.e., $\tanh(W') + W$) introduced in \citet{he2016deep}, and low-rank adaptation methods \cite{jaech2018low,kim2019categorical}, but these did not improve the model performance.

\subsection{Where: Attribute Injection}

\begin{table*}[t]
  \centering
    \begin{tabular}{@{~}lcrrrrrrr@{~}}
    \thickhline
    Datasets & C     & \#train & \#dev & \#test & \#users & \#products & \#docs/user & \#docs/product \\
    \hline
    IMDB  & 10    &     67,426  &     8,381  &     9,112  &   1,310  &   1,635  &          64.82  &          51.94  \\
    Yelp 2013 & 5     &     62,522  &     7,773  &     8,671  &   1,631  &   1,633  &          48.42  &          48.36  \\
    Yelp 2014 & 5     &   183,019  &   22,745  &   25,399  &   4,818  &   4,194  &          47.97  &          55.11  \\
    \hline
    \end{tabular}%
  \caption{Statistics of the datasets used for the Sentiment Classification task.}
  \label{tab:datasets}%
\end{table*}%

Using the approaches described above, we can inject attribute representation into four different parts of the model. This section describes what it means to inject attributes to a certain location and why previous work have been injecting them in the worst location (i.e., in the attention mechanism).

\paragraph{In the attention mechanism}

Injecting attributes to the attention mechanism means that we bias the selection of more informative words during pooling. For example, in Figure \ref{fig:att}, a user may find delicious drinks to be the most important aspect in a restaurant. Injection in the attention mechanism would bias the selection of words such as \textit{wine}, \textit{smooth}, and \textit{sweet} to create the document encoding. This is the standard location in the model to inject the attributes, and several \cite{chen2016neural,amplayo2018cold} have shown how the injected attention mechanism selects different words when the given user or product is different.

We argue, however, that attention mechanism is not the best location to inject the attributes. This is because we cannot obtain user- or product-biased sentiment information from the representation. In the example above, although we may be able to select, with user bias, the words \textit{wine} and \textit{sweet} in the text, we do not know whether the user has a positive or negative sentiment towards these words (e.g., Does the user like wine? How about \textit{sweet} wines? etc.). In contrast, the three other locations we discuss below use the attributes to modify how the model looks at sentiment at different levels of textual granularity.

\paragraph{In the word embedding}

Injecting attributes to the word embedding means that we bias the sentiment intensity of a word independent from its neighboring context. For example, if a user normally uses the words \textit{tasty} and \textit{delicious} with a less and more positive intensity, respectively, the corresponding attribute-injected word embeddings would come out less similar, despite both words being synonymous.

\paragraph{In the BiLSTM encoder}

Injecting attributes to the encoder means that we bias the contextualization of words based on their neighbors in the text. For example, if a user likes their cake sweet but their drink with no sugar, the attribute-injected encoder would give a positive signal to the encoding of \textit{sweet} in the text ``\textit{the cake was sweet}'' and a negative signal in the text ``\textit{the drink was sweet}''.

\paragraph{In the logistic classifier}

Injecting attributes to the classifier means that we bias the probability distribution of sentiment based on the final document encoding. If a user tends to classify the sentiment of reviews about sweet cakes as highly positive, then the model would give a high probability to highly positive sentiment classes for texts such as ``\textit{the cake was sweet}''.

\section{Experiments}
\label{sec:experiments}

\begin{figure}[t]
    \centering
    \includegraphics[width=0.8\columnwidth]{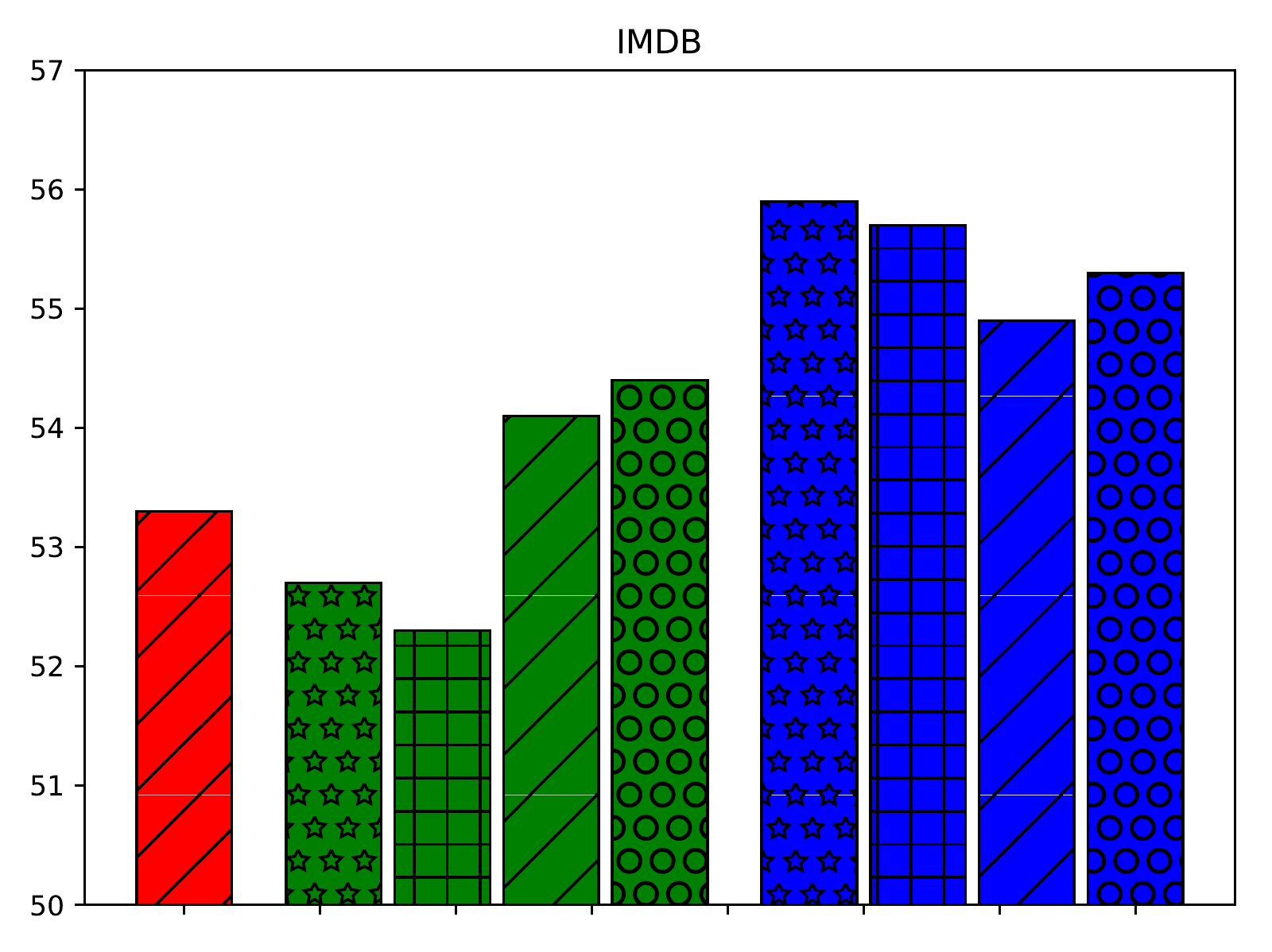}
    \includegraphics[width=0.8\columnwidth]{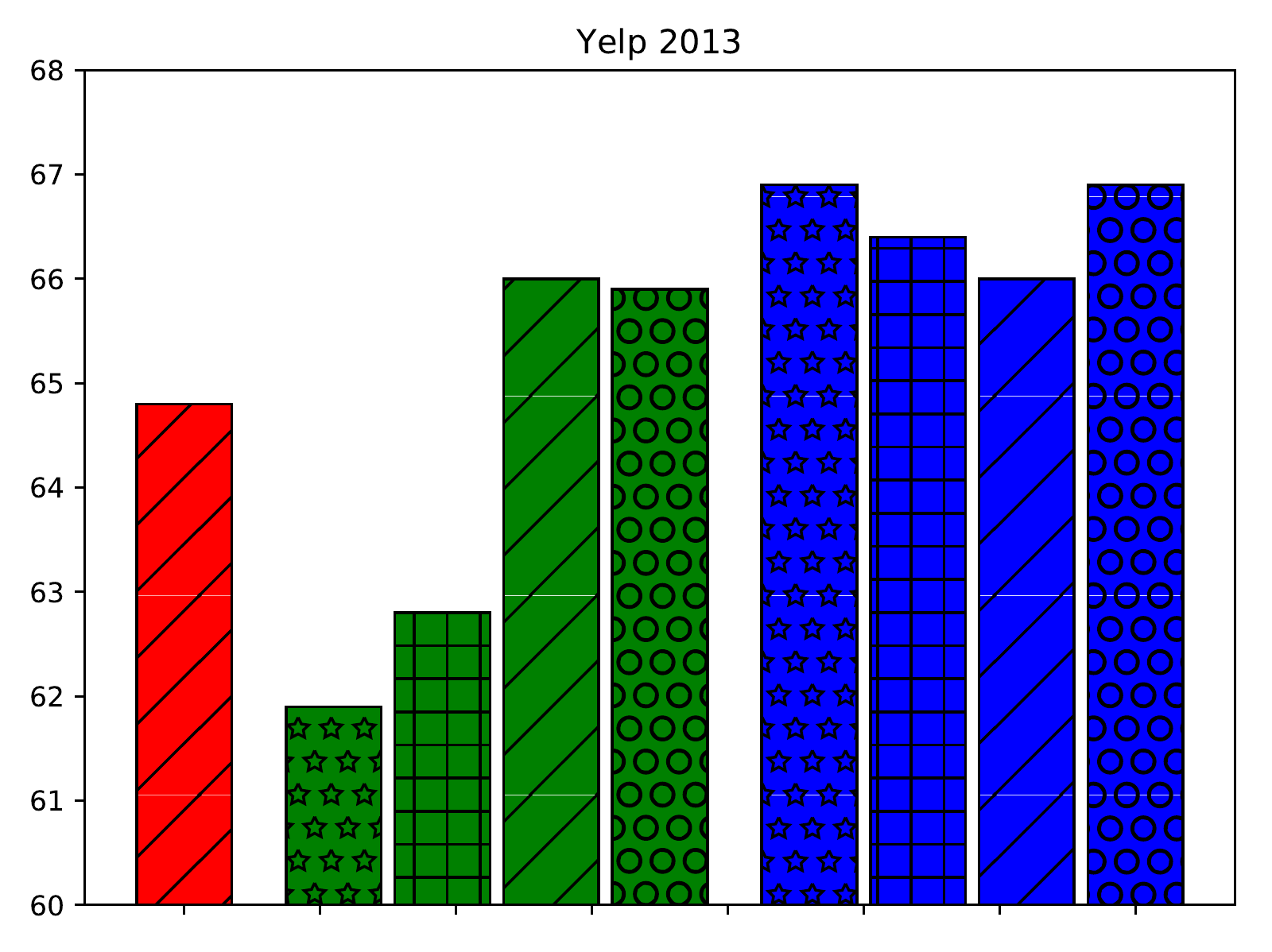}
    \includegraphics[width=0.8\columnwidth]{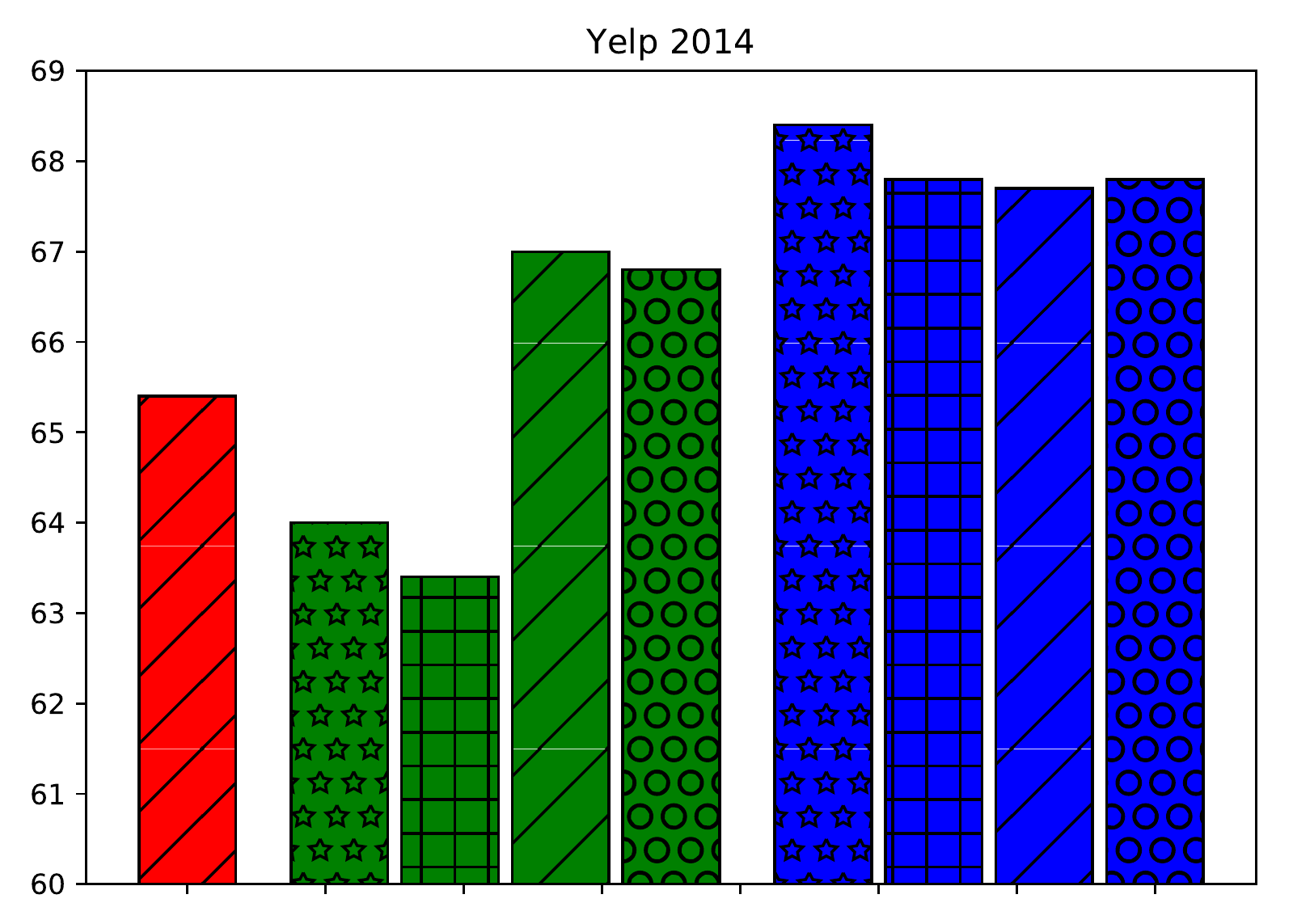}
    \includegraphics[width=0.7\columnwidth]{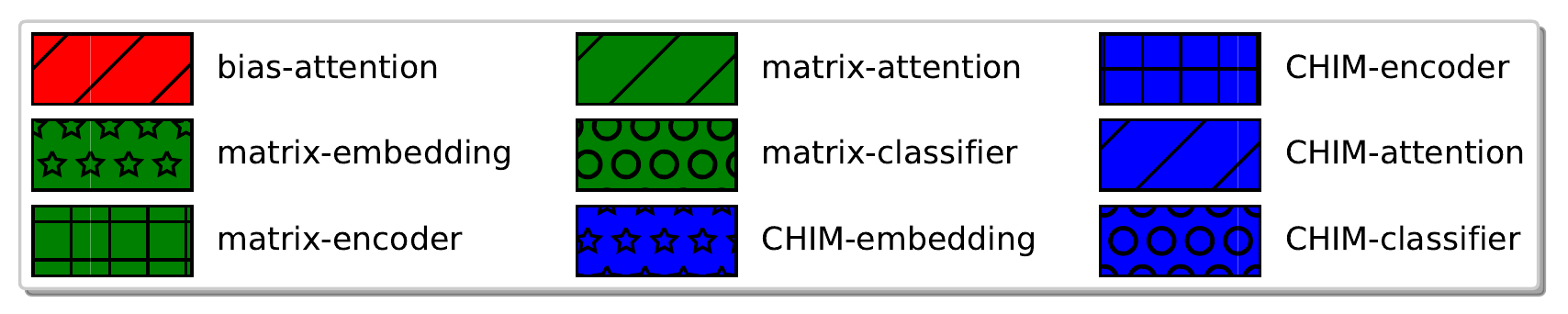}
    \caption{Accuracies (y-axis) of different attribute representation (bias, matrix, CHIM) and injection (emb: embed, enc: encode, att: attend, cls: classify) approaches on the development set of the datasets.}
    \label{fig:dev_set}
\end{figure}

\subsection{General Setup}

We perform experiments on two tasks. The first task is Sentiment Classification, where we are tasked to classify the sentiment of a review text, given additionally the user and product information as attributes. The second task is Attribute Transfer, where we attempt to transfer the attribute encodings learned from the sentiment classification model to solve two other different tasks: (a) Product Category Classification, where we are tasked to classify the category of the product, and (b) Review Headline Generation, where we are tasked to generate the title of the review, given only both the user and product attribute encodings. Datasets, evaluation metrics, and competing models are different for each task and are described in their corresponding sections.

Unless otherwise stated, our models are implemented with the following settings. We set the dimensions of the word, user, and product vectors to 300. We use pre-trained GloVe embeddings\footnote{\url{https://nlp.stanford.edu/projects/glove/}} 
\cite{pennington2014glove} to initialize the word vectors. We also set the dimensions of the hidden state of BiLSTM to 300 (i.e., 150 dimensions for each of the forward/backward hidden state). The chunk size factors $C_1$ and $C_2$ are both set to 15. We use dropout \cite{srivastava2014dropout} on all non-linear connections with a dropout rate of 0.1. We set the batch size to 32. Training is done via stochastic gradient descent over shuffled mini-batches with the Adadelta update rule \cite{zeiler2012adadelta} and with $l_2$ constraint \cite{hinton2012improving} of 3. We perform early stopping using the development set. Training and experiments are done using an NVIDIA GeForce GTX 1080 Ti graphics card.

\subsection{Sentiment Classification}

\paragraph{Datasets and Evaluation}

We use the three widely used sentiment classification datasets with user and product information available: IMDB, Yelp 2013, and Yelp 2014 datasets\footnote{\url{https://drive.google.com/open?id=1PxAkmPLFMnfom46FMMXkHeqIxDbA16oy}}. These datasets are curated by \citet{tang2015learning}, where they ensured twenty-core for both users and products (i.e., users have at least twenty products and vice versa), split them into train, dev, and test sets with an 8:1:1 ratio, and tokenized and sentence-split using the Stanford CoreNLP \cite{manning2014stanford}. 
Dataset statistics are shown in Table \ref{tab:datasets}.
Evaluation is done using two metrics: the accuracy which measures the overall sentiment classification performance, and RMSE which measures the divergence between predicted and ground truth classes.

\paragraph{Comparisons of different attribute representation and injection methods}

To conduct a fair comparison among the different methods described in Section \ref{sec:models}, we compare these methods when applied to our base model using the development set of the datasets. Specifically, we use a smaller version of our base model (with dimensions set to 64) and incorporate the user and product attributes using nine different approaches: (1) \textbf{bias-attention}: the bias-based method injected to the attention mechanism, (2-5) the matrix-based method injected to four different locations (\textbf{matrix-embedding}, \textbf{matrix-encoder}, \textbf{matrix-attention}, \textbf{matrix-classifier}), and (6-9) the CHIM-based method injected to four different locations (\textbf{CHIM-embedding}, \textbf{CHIM-encoder}, \textbf{CHIM-attention}, \textbf{CHIM-classifier}). We then calculate the accuracy of each approach for all datasets.

Results are shown in Figure \ref{fig:dev_set}. The figure shows that bias-attention consistently performs poorly compared to other approaches. As expected, matrix-based representations perform the worst when injected to embeddings and encoder, however we can already see improvements over bias-attention when these representations are injected to attention and classifier. This is because the number of parameters used in the the weight matrices of attention and classifier are relatively smaller compared to those of embeddings and encoder, thus they are easier to optimize. The CHIM-based representations perform the best among other approaches, where CHIM-embedding garners the highest accuracy across datasets. Finally, even when using a better representation method, CHIM-attention consistently performs the worst among CHIM-based representations. This shows that attention mechanism is not the optimal location to inject attributes.

\paragraph{Comparisons with models in the literature}

\begin{table*}[t]
  \centering
    \begin{tabular}{@{~}lllcccccc@{~}}
    \thickhline
    \multicolumn{3}{c}{Model} & \multicolumn{2}{c}{IMDB} & \multicolumn{2}{c}{Yelp 2013} & \multicolumn{2}{c}{Yelp 2014} \\
    \multicolumn{1}{c}{Name} & \multicolumn{1}{c}{Base Model} & \multicolumn{1}{c}{Injection} & Acc   & RMSE  & Acc   & RMSE  & Acc   & RMSE \\
    \hline
    UPNN  & CNN   & embedding, classifier & 43.5  & 1.602 & 59.6  & 0.748 & 60.8  & 0.764 \\
    UPDMN & LSTM  & memory networks & 46.5  & 1.351 & 63.9  & 0.662 & 61.3  & 0.720 \\
    NSC   & HierLSTM & attention & 53.3  & 1.281 & 65.0  & 0.692 & 66.7  & 0.654 \\
    DUPMN & HierLSTM & memory networks & 53.9  & 1.279 & 66.2  & 0.667 & \underline{67.6}  & 0.639 \\
    PMA   & HierLSTM & attention$^1$ & 54.0  & 1.301 & 65.8  & 0.668 & 67.5  & 0.641 \\
    HCSC  & BiLSTM+CNN & attention$^2$ & \underline{54.2}  & 1.213 & 65.7  & \underline{0.660} & --     & -- \\
    CMA   & LSTM+HierAtt & attention$^3$ & 54.0  & \underline{1.191} & \underline{66.4}  & 0.677 & \underline{67.6}  & \underline{0.637} \\
    \hline
    \multirow{4}[0]{*}{\makecell{CHIM\\(Ours)}} & BiLSTM & embedding & \textbf{\underline{56.4}} & \textbf{\underline{1.161}} & \textbf{\underline{67.8}} & 0.646 & \textbf{\underline{69.2}} & 0.629 \\
          & BiLSTM & encoder & 55.9  & 1.234 & 67.0  & 0.659 & 68.4  & 0.631 \\
          & BiLSTM & attention & 54.4  & 1.219 & 66.5  & 0.664 & 68.5  & 0.634 \\
          & BiLSTM & classifier & 55.5  & 1.219 & 67.5  & \textbf{\underline{0.641}} & 68.9  & \textbf{\underline{0.622}} \\
    \thickhline
    \end{tabular}%
  \caption{Sentiment classification results of competing models based on accuracy and RMSE metrics on the three datasets. \underline{Underlined} values correspond to the best values for each block. \textbf{Boldfaced} values correspond to the best values across the board. $^1$uses additional external features, $^2$uses a method that considers cold-start entities, $^3$uses separate bias-attention for user and product.}
  \label{tab:sentiment_results}%
\end{table*}%

We also compare with models from previous work, listed below: 

\begin{enumerate}
    \item \textbf{UPNN} \cite{tang2015learning} uses a CNN classifier as base model and incorporates attributes as user- and product-specific weight parameters in the word embeddings and logistic classifier.
    \item \textbf{UPDMN} \cite{dou2017capturing} uses an LSTM classifier as base model and incorporates attributes as a separate deep memory network that uses other related documents as memory.
    \item \textbf{NSC} \cite{chen2016neural} uses a hierarchical LSTM classifier as base model and incorporates attributes using the bias-attention method on both word- and sentence-level LSTMs.
    \item \textbf{DUPMN} \cite{long2018dual} also uses a hierarchical LSTM as base model and incorporates attributes as two separate deep memory network, one for each attribute.
    \item \textbf{PMA} \cite{zhu2017parallel} is similar to NSC but uses external features such as the ranking preference method of a specific user.
    \item \textbf{HCSC} \cite{amplayo2018cold} uses a combination of BiLSTM and CNN as base model, incorporates attributes using the bias-attention method, and also considers the existence of cold start entities.
    \item \textbf{CMA} \cite{ma2017cascading} uses a combination of LSTM and hierarchical attention classifier as base model, incorporates attributes using the bias-attention method, and does this separately for user and product.
\end{enumerate}

Notice that most of these models, especially the later ones, use the bias-attention method to represent and inject attributes, but also employ a more complex model architecture to enjoy a boost in performance.

Results are summarized in Table \ref{tab:sentiment_results}. On all three datasets, our best results outperform all previous models based on accuracy and RMSE. Among our four models, CHIM-embedding performs the best in terms of accuracy, with performance increases of 2.4\%, 1.3\%, and 1.6\% on IMDB, Yelp 2013, and Yelp 2014, respectively. CHIM-classifier performs the best in terms of RMSE, outperforming all other models on both Yelp 2013 and 2014 datasets. Among our models, CHIM-attention mechanism performs the worst, which shows similar results to our previous experiment (see Figure \ref{fig:dev_set}).
We emphasize that our models use a simple BiLSTM as base model, and extensions to the base model (e.g., using multiple hierarchical LSTMs as in \citealt{wu2018improving}), as well as to other aspects (e.g., consideration of cold-start entities as in \citealt{amplayo2018cold}), are orthogonal to our proposed attribute representation and injection method. Thus, we expect a further increase in performance when these extensions are done.

\subsection{Attribute Transfer}

In this section, we investigate whether it is possible to transfer the attribute encodings, learned from the sentiment classification model, to other tasks: product category classification and review headline generation.
The experimental setup is as follows. First, we train a sentiment classification model using an attribute representation and injection method of choice to learn the attribute encodings. Then, we use these fixed encodings as input to the task-specific model.

\paragraph{Dataset}

We collected a new dataset from Amazon\footnote{\url{https://s3.amazonaws.com/amazon-reviews-pds/tsv/amazon_reviews_multilingual_US_v1_00.tsv.gz}}, which includes the product category and the review headline, aside from the review text, the sentiment score, and the user and product attributes. Following \citet{tang2015learning}, we ensured that both users and products are twenty-core, split them into train, dev, and test sets with an 8:1:1 ratio, and tokenized and sentence-split the text using Stanford CoreNLP \cite{manning2014stanford}. The final dataset contains 77,028 data points, with 1,728 users and 1,890 products. This is used as the sentiment classification dataset.

To create the task-specific datasets, we split the dataset again such that no users and no products are seen in at least two different splits. That is, if user $u$ is found in the train set, then it should not be found in the dev and the test sets. We remove the user-product pairs that do not satistfy this condition. We then append the corresponding product category and review headline for each user-product pair. The final split contains 46,151 training, 711 development, and 840 test instances.
It also contains two product categories: Music and Video DVD. The review headline is tokenized using SentencePiece\footnote{\url{https://github.com/google/sentencepiece}} with 10k vocabulary. The datasets are released here for reproducibility: \url{https://github.com/rktamplayo/CHIM}.

\paragraph{Evaluation}

In this experiment, we compare five different attribute representation and injection methods: (1) the bias-attention method, and (2-5) the CHIM-based representation method injected to all four different locations in the model. We use the attribute encodings, which are learned from pre-training on the sentiment classification dataset, as input to the transfer tasks, in which they are fixed and not updated during training. As a baseline, we also show results when using encodings of randomly set weights. Moreover, we additionally show the majority class as additional baseline for product category classification.

For the product category classification task, we use a logistic classifier as the classification model and accuracy as the evaluation metric. For the review headline generation task, we use an LSTM decoder as the generation model and perplexity as the evaluation metric.

\paragraph{Results}

For the product category classification task, the results are reported in Table \ref{tab:transfer_results}. The table shows that representations learned from CHIM-based methods perform better than the random baseline. The best model, CHIM-encoder, achieves an increase of at least 3 points in accuracy compared to the baseline. This means that, interestingly, CHIM-based attribute representations have also learned information about the category of the product. In contrast, representations learned from the bias-attention method are not able to transfer well on this task, leading to worse results compared to the random and majority baseline. Moreover, CHIM-attention performs the worst among CHIM-based models, which further shows the ineffectiveness of injecting attributes to the attention mechanism.

\begin{table}[t]
  \centering
    \begin{tabular}{@{~}lcc @{~}}
    \thickhline
    Method & Accuracy & Perplexity \\
    \hline
    Majority & 60.12 & -- \\
    Random & 60.67 $\pm$ 0.27 & 43.53 \\
    Bias - Attention & \textcolor{red}{58.74 $\pm$ 0.49} & \textcolor{red}{44.00} \\
    CHIM - Embedding & 62.26 $\pm$ 0.22 & 42.71\\
    CHIM - Encoder & 64.62 $\pm$ 0.34 & 42.65\\
    CHIM - Attention & 60.95 $\pm$ 0.15 & 42.78\\
    CHIM - Classifier & 61.83 $\pm$ 0.43 & 42.69\\
    \thickhline
    \end{tabular}%
  \caption{Accuracy (higher is better) and perplexity (lower is better) of competing models on the Amazon dataset for the transfer tasks on product category classification and review headline generation, respectively. Accuracy intervals are calculated by running the model 10 times. Performance worse than the random and majority baselines are colored red.}
  \label{tab:transfer_results}%
\end{table}%

Results for the review headline generation task are also shown in Table \ref{tab:transfer_results}. The table shows less promising results, where the best model, CHIM-encoder, achieves a decrease of 0.88 points in perplexity from the random encodings.
Although this still means that some information has been transferred, one may argue that the gain is too small to be considered significant.
However, it has been well perceived, that using only the user and product attributes to generate text is unreasonable, since we expect the model to generate coherent texts using only two vectors. This impossibility is also reported by \citet{dong2017learning} where they also used sentiment information, and \citet{ni2018personalized} where they additionally used learned aspects and a short version of the text to be able to generate well-formed texts.
Nevertheless, the results in this experiment agree to the results above regarding injecting attributes to the attention mechanism; bias-attention performs worse than the random baseline, and CHIM-attention performs the worst among CHIM-based models.

\subsection{Where should attributes be injected?}

All our experiments unanimously show that (a) the bias-based attribute representation method is not the most optimal method, and (b) injecting attributes in the attention mechanism results to the worst performance among all locations in the model, regardless of the representation method used. The question ``where is the best location to inject attributes?'' remains unanswered, since different tasks and settings produce different best models. That is, CHIM-embedding achieves the best accuracy while CHIM-classifier achieves the best RMSE on sentiment classification. Moreover, CHIM-encoder produces the most transferable attribute encoding for both product category classification and review headline generation. The suggestion then is to conduct experiments on all locations and check which one is best for the task at hand.

\begin{figure}[t]
    \centering
    \includegraphics[width=\columnwidth]{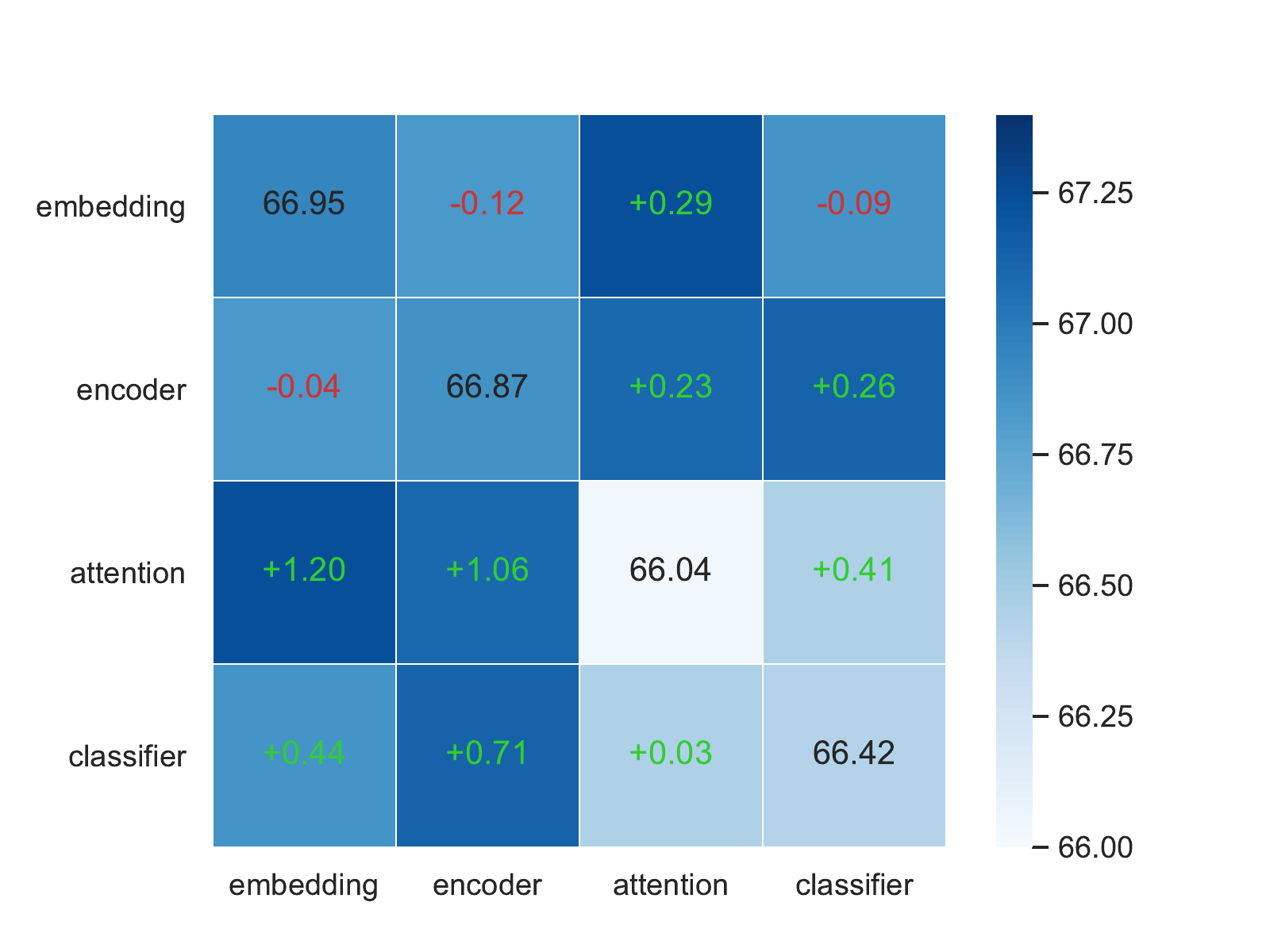}
    \caption{Heatmap of the accuracies of singly and jointly injected CHIM models. Values on each cell represents either the accuracy (for singly injected models) or the difference between the singly and doubly injected models per row.}
    \label{fig:joint_inject}
\end{figure}

Finally, we also investigate whether injecting in to more than one location would result to better performance. Specifically, we jointly inject in two different locations at once using CHIM, and do this for all possible pairs of locations. We use the smaller version of our base model and calculate the accuracies of different models using the development set of the Yelp 2013 dataset.
Figure \ref{fig:joint_inject} shows a heatmap of the accuracies of jointly injected models, as well as singly injected models. Overall, the results are mixed and can be summarized into two statements. Firstly, injecting on the embedding and another location (aside from the attention mechanism) leads to a slight decrease in performance. Secondly and interestingly, injecting on the attention mechanism and another location always leads to the highest increase in performance, where CHIM-attention+embedding performs the best, outperforming CHIM-embedding. This shows that injecting in different locations might capture different information, and we leave this investigation for future work.

\section{Related Work}

\subsection{Attributes for Sentiment Classification}

Aside from user and product information, other attributes have been used for sentiment classification. Location-based \cite{yang2017identifying} and time-based \cite{fukuhara2007understanding} attributes help contextualize the sentiment geographically and temporally. Latent attributes that are learned from another model have also been employed as additional features, such as latent topics from a topic model \cite{lin2009joint}, latent aspects from an aspect extraction model \cite{jo2011aspect}, argumentation features  \cite{wachsmuth2015sentiment}, among others. Unfortunately, current benchmark datasets do not include these attributes, thus it is practically impossible to compare and use these attributes in our experiments.
Nevertheless, the methods in this paper are not limited to only user and product attributes, but also to these other attributes as well, whenever available.

\subsection{User/Product Attributes for NLP Tasks}

Incorporating user and product attributes to NLP models makes them more personalized and thus user satisfaction can be increased \cite{baruzzo2009general}. Examples of other NLP tasks that use these attributes are text classification \cite{kim2019categorical}, language modeling \cite{jaech2018low}, text generation \cite{dong2017learning,ni2018personalized}, review summarization \cite{yang2018personalized}, machine translation \cite{michel2018extreme}, and dialogue response generation \cite{zhang2017neural}. On these tasks, the usage of the bias-attention method is frequent since it is trivially easy and there have been no attempts to investigate different possible methods for attribute representation and injection. We expect this paper to serve as the first investigatory paper that contradicts to the positive results previous work have seen from the bias-attention method.

\section{Conclusions}

We showed that the current accepted standard for attribute representation and injection, i.e.  bias-attention, which incorporates attributes as additional biases in the attention mechanism, is the least effective method. We proposed to represent attributes as chunk-wise importance weight matrices (CHIM) and showed that this representation method significantly outperforms the bias-attention method. Despite using a simple BiLSTM classifier as base model, CHIM significantly outperforms the current state-of-the-art models, even when those models use a more complex base model architecture. 
Furthermore, we conducted several experiments that conclude that injection to the attention mechanism, no matter which representation method is used, garners the worst performance. This result contradicts previously reported conclusions regarding attribute injection to the attention mechanism.
Finally, we show promising results on transferring the attribute representations from sentiment classification, and use them to two different tasks such as product category classification and review headline generation.

\section*{Acknowledgments}

We would like to thank the anonymous reviewers for their helpful feedback and suggestions. Reinald Kim Amplayo is grateful to be supported by a Google PhD Fellowship.

\bibliography{emnlp-ijcnlp-2019}
\bibliographystyle{acl_natbib}

\end{document}